\documentclass[sigconf]{acmart}
\usepackage{xcolor}
\hypersetup{colorlinks=true}

\usepackage{mdframed}
\newmdenv[
  linewidth=0.4pt,
  linecolor=black!50,
  backgroundcolor=black!3,
  skipabove=8pt,
  skipbelow=8pt,
  innerleftmargin=3pt,
  innerrightmargin=3pt
]{questionbox}

\usepackage{amsthm}
\usepackage{stmaryrd}
\usepackage{graphicx}
\usepackage[most]{tcolorbox}
\usepackage{makecell}
\usepackage{multirow}
\usepackage{array}
\usepackage[table]{xcolor}
\usepackage{float}
\newcolumntype{C}[1]{>{\centering\arraybackslash}m{#1}}

\usepackage{microtype}

\theoremstyle{definition}
\newtheorem{definition}{Definition}

\AtBeginDocument{%
  }

\copyrightyear{2026}
\acmYear{2026}
\setcopyright{cc}
\setcctype{by}
\acmConference[KDD '26]{Proceedings of the 32nd ACM SIGKDD Conference on Knowledge Discovery and Data Mining V.2}{August 09--13, 2026}{Jeju Island, Republic of Korea}
\acmBooktitle{Proceedings of the 32nd ACM SIGKDD Conference on Knowledge Discovery and Data Mining V.2 (KDD '26), August 09--13, 2026, Jeju Island, Republic of Korea}
\acmDOI{10.1145/3770855.3817495}
\acmISBN{979-8-4007-2259-2/2026/08}

\begin{document}

\title{{\textit{In a Streaming World, Should You Stand Still?}} \\ A Comprehensive Benchmark of Anomaly Detection in Streams}


\author{Magali Parrino}
\orcid{0009-0008-6260-0921}
\affiliation{
  \institution{EDF R\&D}
  \city{Chatou}
  \country{France}}
  
\affiliation{%
  \institution{\makebox[0pt]{ENS, PSL University, CNRS, Inria}}
  \city{Paris}
  \country{France}
}
\email{magali.parrino@edf.fr}

\author{Antoine Ajenjo}
\orcid{0000-0002-4003-5334}
\affiliation{%
  \institution{EDF R\&D}
  \city{Chatou}
  \country{France}
}
\email{antoine.ajenjo@edf.fr}

\author{Emmanuel Remy}
\orcid{0000-0003-3423-5919}
\affiliation{%
  \institution{EDF R\&D}
  \city{Chatou}
  \country{France}
}
\email{emmanuel.remy@edf.fr}

\author{Pierre Stephan}
\orcid{0009-0006-6813-4437}
\affiliation{%
  \institution{EDF R\&D}
  \city{Chatou}
  \country{France}
}
\email{pierre.stephan@edf.fr}

\author{Pierre Senellart}
\orcid{0000-0002-7909-5369}
\affiliation{
  \institution{DI~ENS, ENS, PSL University, CNRS, Inria}
  \city{Paris}
  \country{France}
}
\email{pierre.senellart@ens.fr}

\author{Paul Boniol}
\orcid{0000-0001-8516-0123}
\affiliation{%
  \institution{DI~ENS, ENS, PSL University, CNRS, Inria}
  \city{Paris}
  \country{France}
}
\email{paul.boniol@inria.fr}

\renewcommand{\shortauthors}{Magali Parrino et al.} 

\begin{abstract}
Time series anomaly detection (TSAD) is increasingly deployed in streaming settings, where data arrive sequentially and may exhibit non-stationarity. As a result, several works from the recent literature propose streaming anomaly detection methods that rely on incremental updates to adapt over time. However, most of these approaches originate from the streaming outlier detection literature and largely ignore core characteristics of time series anomalies. Moreover, their empirical evaluation is typically conducted on synthetic or small-scale benchmarks with limited diversity, making it unclear whether \textit{streaming} methods are truly advantageous in realistic TSAD scenarios. In this work, we carry out the first large-scale experimental study comparing \textit{streaming} and \textit{static} TSAD methods under a unified streaming evaluation benchmark. We consider a realistic setting in which an initial batch of data is available for model training, followed by online evaluation of both detection accuracy and computational efficiency. In addition, we propose a distribution-drift dataset of real time series, called TSB-\textit{drift}, to isolate scenarios where streaming updates are theoretically justified. Our results show that, contrary to common assumptions, \textbf{\textit{static} TSAD methods significantly outperform streaming approaches in most streaming settings.} Such finding highlights a critical gap between the design of existing streaming methods and the requirements of modern TSAD, and calls for a rethinking of how streaming capabilities should be integrated into TSAD.
\end{abstract}




\maketitle


\newcommand\kddavailabilityurl{https://doi.org/10.5281/zenodo.20310651}

\ifdefempty{\kddavailabilityurl}{}{
\begingroup\small\noindent\raggedright\textbf{KDD Availability Link:}\\
The source code of this paper has been made publicly available at \url{\kddavailabilityurl}.
\endgroup
}

\section{Introduction}

Time series anomaly detection (TSAD) plays a critical role in a wide range of real-world applications, including industrial control~\cite{vibration_monitoring}, energy production~\cite{energy_prod_TS}, healthcare~\cite{MITDB} and Internet-of-Things (IoT) platforms~\cite{NID}. In many of these fields, data are generated continuously and must be processed sequentially, motivating the development of TSAD methods suitable for \textit{Streaming} settings. As a consequence, recent years have seen increasing interest in streaming anomaly detection approaches~\cite{Vazquez23, Cao24, Salles25} that incrementally update their models to handle evolving data distributions~\cite{Gama14}.

A common assumption is that streaming TSAD methods are inherently better suited for streaming data than \textit{static} alternatives. In particular, update mechanisms~\cite{LEAP, MCOD}, forgetting factors~\cite{MemStream, SDOstream}, or incremental model updates~\cite{RSHash, xStream}, are often presented as essential tools for dealing with non-stationarity~\cite{Vazquez23}. However, it remains unclear whether existing streaming methods actually deliver superior performance when applied to realistic TSAD tasks.

This uncertainty lies in the conceptual gap between the literature on streaming anomaly detection and on TSAD. Much of the streaming literature focuses on point-wise outlier detection, where each observation is treated independently and anomalies are defined as individual points~\cite{outlier_stream, SWKNN, MCOD, SDOstream, xStream, RSHash}. In contrast, TSAD fundamentally relies on the temporal structure of the data: anomalies often manifest as subsequences~\cite{Keogh05, Boniol24} or collective patterns that become apparent when contextualized within a broader temporal window. Ignoring these characteristics can lead to well-suited models for detecting isolated outliers but poorly designed to identify anomalies commonly encountered in real-world time series~\cite{WuKeogh22, TSB_UAD}.
 
A second equally important issue concerns evaluation practices. Most streaming anomaly detection methods are evaluated on synthetic datasets~\cite{Cao24, Vazquez23, Ntroumpogiannis23} or on a small number of narrowly scoped real-world time series~\cite{Salles25, Vazquez23}. These benchmarks often lack diversity in terms of domains, temporal characteristics, anomaly types and degrees of non-stationarity. Meanwhile, recent work in TSAD literature has introduced large-scale, heterogeneous benchmarks~\cite{Liu25} that better reflect the variety of real-world applications. To date, however, these modern benchmarks have not been used to assess streaming anomaly detection methods. Therefore, the aforementioned limitations raise a fundamental question: 

\begin{questionbox}
\begin{center}
\textit{Do streaming anomaly detection methods actually perform better than static TSAD methods when evaluated in streaming settings? }    
\end{center}
\end{questionbox}

In this paper, we address this question through our proposed benchmark, called \textbf{\texttt{StrAD}}. 
We evaluate a large collection of static and streaming methods under a unified streaming evaluation framework. 
Our setting assumes an initial batch of data for model training, followed by sequential processing of incoming data. 
We operate in an \textit{unsupervised context}, so the time series may contain anomalies within the training batch. 
Static methods are trained once and remain unchanged, while streaming methods apply their update mechanisms during the online phase. 
Importantly, we define streaming methods not merely as those capable of real-time execution, but as methods that explicitly rely on model updates to improve efficiency or accuracy in the presence of distribution drifts. Static methods used in streaming settings are called \textit{Online} methods.

Our results reveal a striking finding: \textbf{online TSAD methods consistently outperform streaming approaches in most strea\-ming settings, often by a substantial margin}. This performance gap persists even though streaming methods are explicitly designed to adapt over time. Results on \textbf{TSB-\textit{drift}}, our novel distribution-drift subset of TSB-AD-M~\cite{Liu25}, reveal that, although the performance gap narrows in the presence of concept drifts, streaming methods do not manage to reverse the trend. 
These findings challenge prevailing assumptions in the streaming TSAD literature and highlight important limitations of current streaming approaches. Rather than demonstrating an inherent advantage of streaming methods, our study suggests that effective time series modeling and anomaly representation remain more critical than incremental model updates alone. 
Overall, our contributions are as follows:

\begin{itemize}
    \item We first properly define the concept of \textit{Static}, \textit{Online} and \textit{Streaming} TSAD methods (\textbf{Section~\ref{StatOnlStream}}).
    \item We carefully describe our experimental benchmark \textbf{\texttt{StrAD}} and introduce a novel dataset subset of TSB-AD-M, called TSB-\textit{drift} that contains only real-world time series that manifest a concept drift (\textbf{Section~\ref{benchmark}}).
    \item We present our experimental results comparing static, online and streaming TSAD methodologies on both TSB-AD benchmark and TSB-\textit{drift} (\textbf{Section~\ref{experiments}}).
    \item We conclude by discussing the implications of these results for potential future research (\textbf{Section~\ref{discussions}}).
    \item We release our code and results in \textbf{our repository}~\footnote{\href{https://github.com/magaliparrino/StrAD.git}{https://github.com/magaliparrino/StrAD.git}}.
\end{itemize}

\section{Background and Related Work}

We now introduce formal definitions relevant for this paper.

\noindent A \textbf{time series} $T \in \mathbb{R}^n $ is a sequence of values $T_i\in\mathbb{R}$ $[T_0,T_1,...,T_{n-1}]$, where $n=|T|$ is the length of $T$, and $T_i$ is the $i^{th}$ point of $T$. A subsequence of $T$ of length 
$\ell$ starting at position $i$ is defined as
$ T_{i,\ell} = [T_{i}, T_{i+1}, \dots, T_{i+\ell-1}]$ with $i \in [0, n-\ell]$.

\noindent A \textbf{multivariate time series} is a collection of $D$ univariate time series of equal 
length $n$. Formally, we denote $\mathbf{T} \in \mathbb{R}^{(D,n)}$ as $\mathbf{T} = [T^{(0)}, T^{(1)}, \dots, T^{(D-1)}]$,
where for each dimension $j \in [0, D-1]$, the corresponding univariate time series is
$T^{(j)} = [T^{(j)}_{0}, T^{(j)}_{1}, \dots, T^{(j)}_{n-1}]$,
with $T^{(j)}_{i} \in \mathbb{R}$ for $i \in [0, n-1]$.
Moreover, a subsequence of dimension $j$ of length $\ell$ starting at position $i$ is defined as
$T^{(j)}_{i,\ell}$.
The corresponding multivariate subsequence of length $\ell$ starting at position $i$ is then
$\mathbf{T_{i,\ell}} = [T^{(0)}_{i,\ell}, T^{(1)}_{i,\ell}, \dots, T^{(D-1)}_{i,\ell}]$.

\noindent The output of an anomaly detection algorithm applied to a time series $\mathbf{T}$ is commonly expressed as an \textbf{anomaly score sequence}
$S_\mathbf{T} = [S_{\mathbf{T},0}, S_{\mathbf{T},1}, \dots, S_{\mathbf{T},n-1}]$, where $S_{\mathbf{T},i} \in \mathbb{R}$ denotes the anomaly score associated with point $T_i$. Higher scores typically indicate a higher degree of abnormality.

\noindent When available, the ground-truth annotation is represented by a \textbf{label sequence}
$Y = [y_0, y_1, \dots, y_{n-1}]$, where $y_i \in \{0,1\}$ indicates whether point $T_i$ is normal ($y_i = 0$) or anomalous ($y_i = 1$). These labels are used exclusively for evaluation purposes and are not accessible to the detection algorithms during training or inference when in unsupervised settings.

\noindent Anomaly detection in time series is predominantly studied as an \textbf{unsupervised} learning problem, as prior information about anomalies or even normal behavior is often unavailable in real-world scenarios. Moreover, most practical applications involve \textbf{multivariate time series}. Consequently, throughout the remainder of this paper, we restrict our study to \textbf{unsupervised anomaly detection methods} operating on \textbf{multivariate time series}.

\begin{figure*}[tb] 
    \centering
    \includegraphics[width=\linewidth]{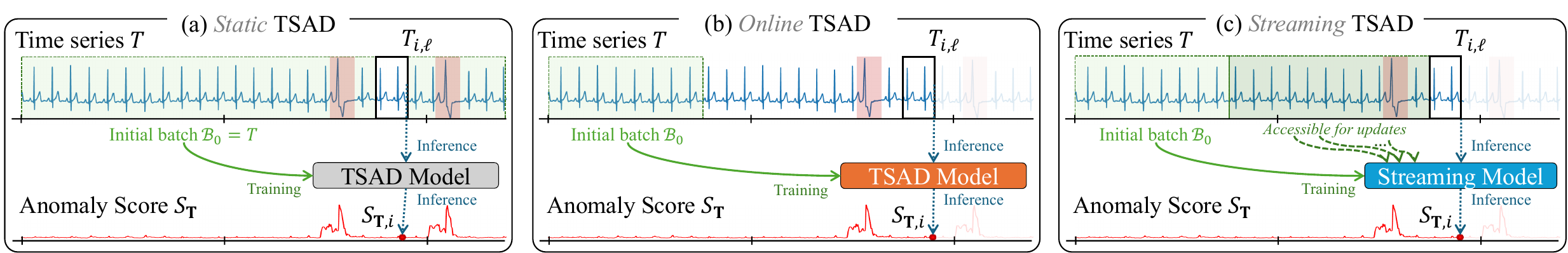}
    \vspace{-0.5cm}
    \caption{\textit{Static}, \textit{Online}, and \textit{Streaming} Time Series Anomaly Detection}
    \vspace{-0.4cm}
    \label{fig:probdef}
\end{figure*}

\subsection{Time Series Anomaly Detection (TSAD)}
\label{TSADRW}

A large panel of TSAD methods have been proposed in the literature~\cite{Boniol24}. The latter can be grouped into three main categories, (i) \textbf{distance-based}, (ii) \textbf{density-based}, and (iii) \textbf{prediction-based}.

\subsubsection{\textbf{Distance-based}} These methods utilize distances between subsequences to detect anomalies. We can identify three sub-catego\-ries: (i) \textit{discord-based}, i.e.,  methods 
that focus on the analysis of subsequences for the purpose of detecting anomalies in time series, mainly by employing 
nearest neighbor distances among subsequences~\cite{Yeh16,Senin15,Yankov07};
(ii) \textit{proximity-based}, i.e.,  methods measuring the density of the neighborhood of particular subsequences~\cite{LOF}; and 
(iii) \textit{clustering-based}, i.e., methods using the distance to a cluster partition~\cite{Boniol21,Boniol21sand}.

\subsubsection{\textbf{Density-based}} These methods detect anomalies by evaluating the density of the points or subsequences into a specific representation space. We identify four sub-categories: (i) \textit{distribution-based} such as the HBOS~\cite{HBOS} method, which estimates the distribution of each feature using histograms and detects anomalies as points that fall into low-density regions
; (ii) \textit{graph-based} in which a method, such as Series2Graph~\cite{Boniol20S2G}, converts the time series into graphs to facilitate the detection of anomalies
; (iii) \textit{tree-based} such as Isolation Forest~\cite{IForest} which groups points or subsequences into different trees
; and (iv) \textit{encoding-based} methods, such as PCA that projects data onto a lower dimensional subspace to identify the deviations from the dominant structure.

\subsubsection{\textbf{Prediction-based}} These methods detect anomalies using prediction errors to a given self-supervised task. We can identify two sub-categories: (i) \textit{forecasting-based}, such as recurrent~\cite{LSTMAD} or convolutional neural network~\cite{CNN} that use the past values as input, predict the following one, and use the forecasting error as the anomaly score; and (ii) \textit{reconstruction-based}, such as AutoEncoder approaches~\cite{AutoEncoder} trained to reconstruct the time series and that use the reconstruction error as an anomaly score.

\subsection{Streaming Anomaly Detection}
\label{streamRW}

There are two main challenges when it comes to streaming TSAD: the unknown incoming data, whose distribution can be drastically different from the historical data and may force an update of the model, and the computational constraints: limited memory and short execution time in case of real-time analysis~\cite{LODA, RRCF,  RSHash, HSTree}.

\subsubsection{\textbf{Concept Drift}}

In non-stationary environments, the underlying data distribution can change over time, resulting in what is known as \textit{Concept Drift}. In the seminal work~\cite{Gama14}, it is formally defined as occurring between two time points $t_0$ and $t_1$ if the joint distribution $P_t(X,Y)$ of the input variables $X$ and the labels $Y$ changes, such that: $\exists X : P_{t_0}(X, Y) \neq P_{t_1}(X, Y).$
However , concept drift was initially introduced in the context of \textit{supervised learning}~\cite{Gama14}, where both the labels $Y$ and the data $X$ are available. It enabled the distinction between \textit{virtual drift} (changes in the incoming data distribution $P(X)$) 
and \textit{real concept drift} (changes in $P(Y|X)$).

In \textit{unsupervised settings}, $Y$ is unavailable, hence we cannot directly measure $P(Y|X)$.  We follow the common assumption in streaming literature~\cite{Vazquez23, Cao24} that significant changes in $P(X)$ serve as a proxy for concept drift. A shift in $P(X)$ means that the statistical patterns learned during training no longer match the incoming data. In this paper, we consider a concept drift to occur when the distribution differs between historical and new data.

\subsubsection{\textbf{Update Mechanism (UM)}}
\label{sec:UM}
To handle concept drift, streaming models contain an update mechanism. 
We identify two categories: (i) \textbf{numerical} and (ii) \textbf{structural} updates. 
In the former, models set their internal geometry during the training phase (using random projections for models such as LODA~\cite{LODA} and xStream~\cite{xStream}, or space partitioning for RSHash~\cite{RSHash}, HSTree~\cite{HSTree} and SDOstream~\cite{SDOstream}). The update will merely consist in adjusting frequency counts or mass profile in these fixed regions. As such, 
HSTree walks down the fixed trees until the new point reaches a leaf and raises the mass of every node on the path. \textbf{Structural updates}, on the other hand, correspond to evolving structures that adapt to concept drifts. RRCF~\cite{RRCF} inserts and deletes nodes for each new arrival while MCOD~\cite{MCOD} manages dynamical clustering within sliding windows. This process, although better for capturing complex shifts, results in higher computational cost.

\subsubsection{\textbf{Memory Management (MM)}}
\label{sec:MM}
Three strategies exist: (i) \textbf{tumbling memory}, (ii) \textbf{sliding memory} and (iii) \textbf{soft forgetting}. The former includes methods that
compare a reference window with a current building window. Once the current window matches the size of the reference window, it is tumbled as the new reference window, and a new current window is built incrementally. The latter is used in xStream~\cite{xStream} and 
HSTree~\cite{HSTree}. In contrast, models employing \textbf{sliding memory} operate under a First In First Out 
paradigm, removing the oldest data when new data are added, whether as individual points (e.g., SWKNN~\cite{SWKNN}, MCOD~\cite{MCOD}, RRCF~\cite{RRCF}, RSHash~\cite{RSHash}) or small batches (e.g., LEAP~\cite{LEAP}). Finally, \textbf{soft forgetting} regroups less abrupt memory management.
SDOstream~\cite{SDOstream} employs an aging mechanism, assigning a decreasing weight to observations so that historical data has a diminishing impact on current evaluations. MemStream~\cite{MemStream} maintains a fixed‑size memory of normal points, adding a new point only if it is sufficiently normal and discarding the oldest entry.

\subsection{\textit{Static} vs. \textit{Online} vs. \textit{Streaming} Methods}
\label{StatOnlStream}

We now distinguish TSAD methods based on data availability for anomaly score computation and model updates ability. The latter constitutes the major difference between (i) \textbf{Static}, (ii) \textbf{Online}, (iii)~\textbf{Streaming} methods, depicted in Figure~\ref{fig:probdef} and defined as follows.

\begin{definition}[Static Method]
A \emph{static} method computes the anomaly score at timestamp $i$ using the entire
multivariate time series $\mathbf{T}$.
Formally, the anomaly score $s_{i}$ at timestamp $i$ may depend on any observation
$T^{(j)}_k$ for all dimensions $j \in [0, D-1]$ and all timestamps $k \in [0, |\mathbf{T}|-1]$.
There is no temporal ordering constraint.
\end{definition}

\begin{definition}[Online Method]
An \emph{online} method assumes access to an initial batch
$\mathcal{B}_0=\mathbf{T}_{0,m}$ for some $m < n$.
After this initial batch, the model parameters are fixed.
For any timestamp $i \ge m$, the anomaly score $s_{i}$ is computed using only $\mathcal{B}_0$ and 
 $T^{(j)}_k$ with $k \in [i - \ell, i]$ and $j \in [0, D-1]$.
No model updates are performed..
\end{definition}

\begin{definition}[Streaming Method]
A \emph{streaming} method processes the time series sequentially and allows model
updates over time.
At timestamp $i$, the anomaly score $s_{i}$ is computed using observations available up
to time $i$, i.e., $T^{(j)}_k$ with $k \le i$ and $j \in [0, D-1]$.
After computing $s_i$, the model may update its internal state or parameters.
\end{definition}

\begin{figure*}[tb] 
    \centering
    \includegraphics[width=\linewidth]{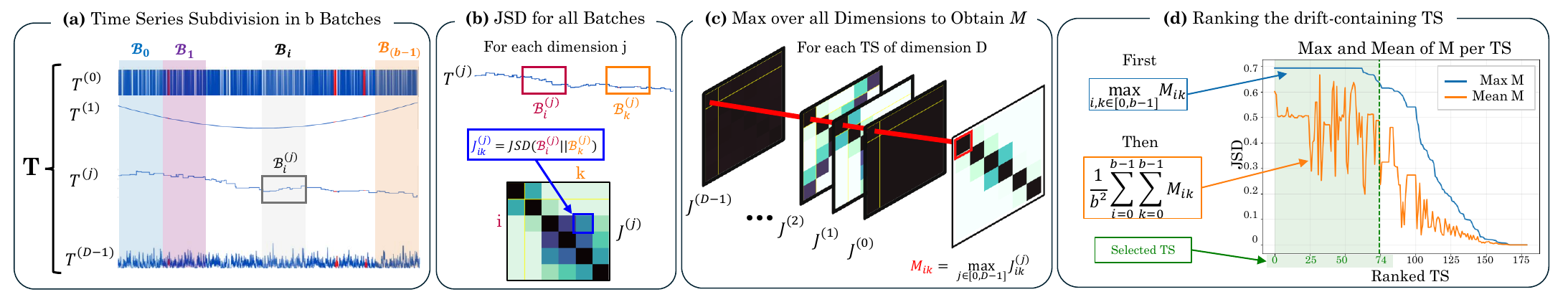}
    \vspace{-0.6cm}
    \caption{Illustration of TSB-\textit{drift} Construction}
    \vspace{-0.3cm}
    \label{fig:schema_drift}
\end{figure*}

These notions are closely related to those introduced in
a recent streaming survey~\cite{Correia24}.
In practice, most methods in TSAD literature (cf. Section~\ref{TSADRW}) are static methods, assuming access to the full time series. Nevertheless, many of these approaches can naturally be deployed in an online setting, by performing training on a initial batch $\mathcal{B}_0$, followed by inference on incoming observations without further model updates. This observation suggests that static-online boundary is 
often methodological rather than intrinsic, and that static formulations do not prevent practical online use.

\subsection{Gap in Existing Evaluations}
\label{gaps}

Despite extensive research on both static and streaming TSAD, current benchmarks exhibit the following three major limitations:

\subsubsection{\textbf{Streams are Time Series}}
Data streams 
consist of ordered observations collected over time. However, the literature on stream outlier detection,
largely treats streams as generic sequences of data points~\cite{Vazquez23}, arguing that computational requirements fundamentally differ. Thus, time-series--specific characteristics are often not explicitly considered. This abstraction has led to a separation between the literature on streaming and TSAD, where TSAD methods are typically overlooked in streaming evaluations.

\subsubsection{\textbf{Anomalies are not Only Points}} Anomalies are not restricted to isolated observations but also manifest as subsequences\\~\cite{Keogh05, Boniol24,Chandola09}, often referred to as collective anomalies~\cite{Chandola09}. 
Detecting such anomalous patterns over contiguous time intervals is a central challenge in the TSAD literature~\cite{Boniol24}. Nevertheless, this notion is sometimes overlooked even within TSAD, especially for multivariate time series, where the complexity of jointly modeling temporal and cross-dimensional structure can lead to a focus on point-wise anomalies. This disconnect is even sharper in streaming outlier literature, which prioritizes identifying 
point anomalies~\cite{outlier_stream, SWKNN, MCOD, SDOstream, xStream, RSHash} over anomalous contiguous sequences.

\subsubsection{\textbf{Is Concept Drift even Real?}}While concept drift is inherent in real-world time series and its treatment is recognized as a central methodological challenge in online and streaming anomaly detection, existing anomaly detection benchmarks provide limited evidence of its occurrence in real time series.
In particular, there is a lack of diverse real datasets with explicitly identified concept drift regarding anomaly detection. Many benchmarks rely on synthetically generated drift patterns \cite{Cao24, Vazquez23}, or employ real-world datasets without conducting a dedicated drift analysis \cite{Vazquez23, Salles25}. As noted by \cite{Cao24}, current streaming datasets typically focus either on anomaly detection or on concept drift, but rarely address both simultaneously. As a result, the practical impact of concept drift on TSAD methods' performance remains difficult to assess under realistic evaluation settings.

\subsection{Objectives and Research Questions}
\label{RQ}

Based on the above-mentioned evaluation gaps for Streaming anomaly detection, the objective of this paper is to propose a comprehensive benchmark and experimental evaluation, aiming to answer the following research questions:

\begin{itemize}
    \item[(Q1)] \textit{Do TSAD methods significantly perform better in static settings rather than in online settings?}
    \item[(Q2)] \textit{Are streaming approaches more relevant than online approaches in streaming settings? }
    \item[(Q3)] \textit{Are streaming approaches more appropriate  than online ones in case of concept drift?}
    \item[(Q4)] \textit{Do streaming methods balance accuracy and efficiency best?}
\end{itemize}

\section{\texttt{StrAD}: Our Proposed Benchmark}
\label{benchmark}

To address these questions, we introduce \texttt{StrAD}, a benchmark designed to enable a systematic evaluation of unsupervised and multivariate anomaly detection methods in streaming settings. \texttt{StrAD} evaluates SOTA streaming models and a broad spectrum of TSAD methods on real-world data.
Overall \texttt{StrAD} contributions are:
\begin{itemize}
\item An evaluation on 17 real-world datasets (c.f. \textbf{Table~\ref{tab:datasets}}).
\item The introduction of TSB-\textit{drift}, a carefully designed collection of time series with 
clear concept drifts  (c.f. \textbf{Table~\ref{tab:datasets}}).
\item 19 static and online TSAD methods (c.f. \textbf{Table~\ref{tab:TSADmethods}}).
\item 10 streaming anomaly detection methods  (c.f. \textbf{Table~\ref{tab:StreamingMethods}}).
\end{itemize}

\begin{table}[tb]
\centering
\caption{Datasets in \textbf{\texttt{StrAD}}. \textit{Drift}: \# of TS 
with 
Concept Drift, \% of Dimensions with Drifts (ratio), and Type}
\vspace{-0.3cm}
\label{tab:datasets}
\resizebox{\columnwidth}{!}{
\begin{tabular}{l|ccc|ccc}
\toprule
\multicolumn{4}{c}{\textit{General}} & \multicolumn{3}{|c}{\textit{Drift}} \\
\midrule
\textbf{Dataset} & \textbf{Category (Field)} & \textbf{\# TS} & \makecell{\textbf{$\mu(D)$}} & \makecell{\textbf{\# TS}} & \makecell{\textbf{ratio}} & \makecell{\textbf{Type}} \\
\midrule
\textbf{Genesis} & Sensor (Robotics) & 1 & 18 & \textcolor{gray}{0/1} & \textcolor{gray}{0\%} & \textcolor{gray}{$\emptyset$}\\
\textbf{MITDB} & Medical & 13 & 13 & \textcolor{gray}{0/13} & \textcolor{gray}{0\%} & \textcolor{gray}{$\emptyset$} \\
\textbf{PSM} & Facility & 1 & 25  & \textcolor{gray}{0/1}& \textcolor{gray}{0\%}  & \textcolor{gray}{$\emptyset$}\\
\textbf{SVDB} & Medical & 31 & 2  & \textcolor{gray}{0/31} & \textcolor{gray}{0\%} & \textcolor{gray}{$\emptyset$} \\
\textbf{MSL} & Sensor (Aerospace) & 16 & 16  &\textcolor{gray}{0/16} & \textcolor{gray}{0\%} & \textcolor{gray}{$\emptyset$} \\
\hline
\rowcolor{gray!20}
\multicolumn{7}{c}{represented in \textbf{TSB-\textit{drift}}} \\
\textbf{Daphnet} & Human Activity & 1 & 9 & 1/1 & 11\% & $\{CP\}$\\
\textbf{GHL} & Sensor (Industry) & 25 & 19 & 23/25 & 15\% & $\{C,CP,RW\}$\\
\textbf{SMD} & Facility & 22 & 38 & 15/22 & 13\% & $\{C,CP,P,RW\}$\\
\textbf{LTDB} & Medical & 5 & 2 & 1/5 & 67\% & $\{P\}$\\
\textbf{TAO} & Environment & 13 & 3 & 8/13 & 54\%  & $\{C\}$\\
\textbf{OPP}. &  \makecell{Human Activity} & 8 & 248 & 7/8 & 23\% & $\{CP, RW\}$\\
\textbf{CreditCard} & Finance & 1 & 29 & 1/1 & 3\% & $\{CP\}$\\
\textbf{CATSv2} & Sensor (Dynamic System) & 6 & 17 & 5/6 & 20\% & $\{C,P,RW\}$\\
\textbf{SMAP} & Sensor (Telemetry) & 27 & 25 & 5/27 & 4\% & $\{C,CP\}$\\
\textbf{SWaT} & Sensor (Cybersecurity) & 2 & 59 & 1/2 & 2\% & $\{C,CP\}$\\
\textbf{GECCO} & Sensor (Water quality) & 1 & 9 & 1/1 & 11\% & $\{C\}$\\
\textbf{Exathlon} & Facility & 27 & 21 & 5/27 & 5\% & $\{CP,RW\}$\\
\bottomrule
\end{tabular}
}
\end{table}

\subsection{TSB-\textit{drift}: a Concept-Drift Subset}

To the best of our knowledge, 
no current streaming benchmark explicitly evaluates performance of anomaly detection on real-world data with identified concept drift.
To address this gap, we introduce \textbf{TSB-\textit{drift}}, a curated subset of real-world time series exhibiting clear concept drifts. This subset is constructed from the TSB-AD benchmark \cite{Liu25}.

\subsubsection{\textbf{TSB-AD-M as a starting point}}

TSB-AD-M consists in 200 multivariate real-world time series from 17 different datasets as presented in Table \ref{tab:datasets}. Those datasets are recurrent in TSAD \cite{Wagner23, Audibert22, Sylligardos25, WuKeogh22} and were selected to avoid identified flaws in widely used datasets~\cite{WuKeogh22, Liu25, Wagner23}. TSB-AD-M is divided into two subsets: \textbf{TSB-AD-M-Tuning}, which contains 20 time series dedicated to hyperparameter tuning, and \textbf{TSB-AD-M-Eval}, which includes the remaining 180 time series used for performance evaluation. Each time series is accompanied by a predefined training segment, ensuring reproducibility and fair comparison across methods. We use the latter as the initial batch $\mathcal{B}_0$ for online and streaming methods.

\subsubsection{\textbf{Building TSB-\textit{drift}}}

We derive TSB-\textit{drift} from TSB-AD-M by quantifying distributional changes within each series. The goal is to identify time series with significant changes in \textit{at least one dimension}, ensuring a high-confidence set of drifting series.

\noindent\textbf{(Step a): Batch subdivision.} 
Each multivariate time series $\mathbf{T}$ is subdivided into consecutive batches of equal size, as illustrated in Figure~\ref{fig:schema_drift}(a). The batch size is the training size $tr$, so that for the $j$-th dimension: $\mathcal{B}_i^{(j)} = T_{i \times tr, tr}^{(j)}$, with $i \in [0, \frac{n}{tr}-1].$
This choice retains the training batch as a historical reference in a streaming context. It balances computational efficiency with the need to preserve the training batch as a historical reference in a streaming context.

\noindent\textbf{(Step b): Measuring distributional change.} 
To quantify distribution changes in the data distribution across batches, we use the Jensen-Shannon Divergence $JSD$ which 
is a symmetric and bounded variant of the Kullback-Leibler divergence $D_{\mathrm{KL}}$. Formally,  let $\mathcal{B}_i^{(j)}$ and $\mathcal{B}_k^{(j)}$ be two batches. We estimate their respective empirical distributions $P_{\mathcal{B}_i^{(j)}}$ and $P_{\mathcal{B}_k^{(j)}}$ by binning the data over the support $\mathcal{X}$ and obtain as follows: 
\begin{equation}
\footnotesize
D_{\mathrm{KL}}({\mathcal{B}_i}^{(j)} \,\|\, {\mathcal{B}_k}^{(j)}) = \sum_{x \in \mathcal{X}} P_{\mathcal{B}_i^{(j)}}(x) \log \left(P_{\mathcal{B}_i^{(j)}}(x)/P_{\mathcal{B}_k^{(j)}}(x) \right)
\end{equation}
\begin{equation}
\footnotesize
J_{ik}^{(j)} = JSD({\mathcal{B}_i}^{(j)} \,\|\, {\mathcal{B}_k}^{(j)}) = \tfrac{1}{2} D_{\mathrm{KL}}({\mathcal{B}_i}^{(j)} \,\|\, A_{ik}^{(j)}) + \tfrac{1}{2} D_{\mathrm{KL}}({\mathcal{B}_k}^{(j)} \,\|\, A_{ik}^{(j)})
\end{equation}
In the above equation, we have $A_{ik}^{(j)} = \frac{1}{2}({\mathcal{B}_i}^{(j)}+{\mathcal{B}_k}^{(j)})$. By computing $J_{ik}^{(j)}$ for every pair of batches $({\mathcal{B}_i}^{(j)}, {\mathcal{B}_k}^{(j)})$  
across all dimensions 
(Figure~\ref{fig:schema_drift}(b)),  we obtain $D$ divergence matrices $J^{(j)}$. 

\noindent\textbf{(Step c): Aggregating across dimensions.} 
The divergence information across dimensions is aggregated into a single drift matrix~$M$, where each cell is
$M_{i,k} = \max_{j} J_{ik}^{(j)}$ as shown in Figure~\ref{fig:schema_drift}(c). This max-pooling approach ensures that a significant drift occurring even in a single dimension is captured.

\noindent\textbf{(Step d): Selecting series with strong drift.} 
We finally rank series primarily by the maximum of $M$ and secondarily by the global mean of $M$ to favor long-lasting drifts. As shown in Figure~\ref{fig:schema_drift}(d), a subset of series displays substantial drifts. We retain the top 75 time series, forming a high-confidence set of series with significant drifts.

\subsubsection{\textbf{Drift Characterisation}}
\label{Driftcarac}

The matrices $J^{(j)}$ obtained for each dimension of the time series allow us to identify recurrent drift patterns through the database as theoretically defined in~\cite{Gama14, Vazquez23}. As illustrated in Table~\ref{tab:drift_matrix_ts}, diagonal heatmaps correspond to \textbf{continuous drifts (C)}, block matrices to \textbf{change points (CP)} and cobbled heatmaps to \textbf{periodic drifts (P)}. We name \textbf{random-walks (RW)} the drifts where we could find no distinctive patterns. Finally, dark heatmaps show the absence of drift.

\subsubsection{\textbf{Synthetic Validation}}

We validate our methodology using a \textbf{controlled synthetic suite}. We create multivariate time series containing each identified drift patterns, as well as a No Drift baseline. We also implement a Virtual Drift scenario to ensure that our approach distinguishes real concept drifts from simple noise fluctuations. More details are available in Appendix~\ref{synthetic_drift}. Our methodology successfully detect all drifts. Moreover, it correctly categorize the Virtual Drift as a low magnitude shift ($max M \approx 0.3$) that does not reach our selection values ($max M \geq 0.65$).

\begin{table}[tb]
\centering
\caption{Characteristic Patterns Throughout TSB-\textit{drift} (Training Batch $\mathcal{B}_0$ is Highlighted in Green)}
\vspace{-0.3cm}
\label{tab:drift_matrix_ts}
\resizebox{\columnwidth}{!}{
\begin{tabular}{
|C{1.4cm}|
C{8.8cm}|
C{1.7cm}|
C{1.8cm}|
C{1.6cm}|}
\hline
\makecell{\textbf{Matrix} \\ \textbf{Pattern}} & \textbf{Time Series} & \makecell{\textbf{Concept} \\ \textbf{Drift}} & \makecell{\textbf{ Dataset} \\ (id TS, d)} & \makecell{\textbf{Occurrence} \\ \textbf{Examples}}\\
\hline
\centering\includegraphics[width=\linewidth]{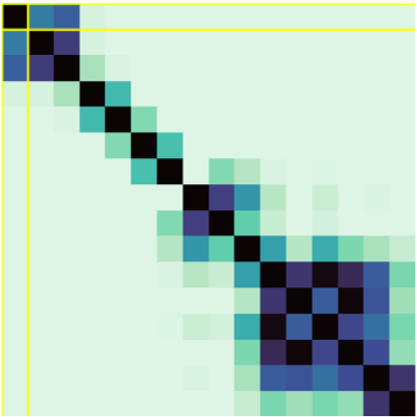} & 
\centering\includegraphics[width=\linewidth]{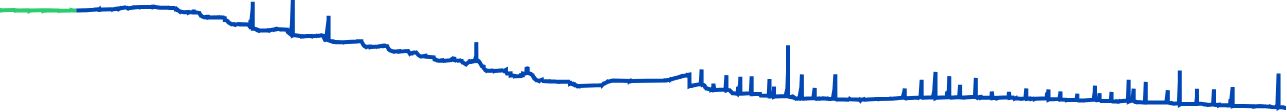}&
\makecell{Continuous \\ \textbf{(C)}} & \makecell{\textbf{SWaT} \\ \texttt{\textbf{id}: 2} \\ \texttt{\textbf{dim}: AIT201}} & \makecell{Sensor\\Degradation} \\
\hline
\centering\includegraphics[width=\linewidth]{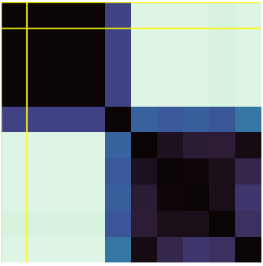}& 
\centering\includegraphics[width=\linewidth]{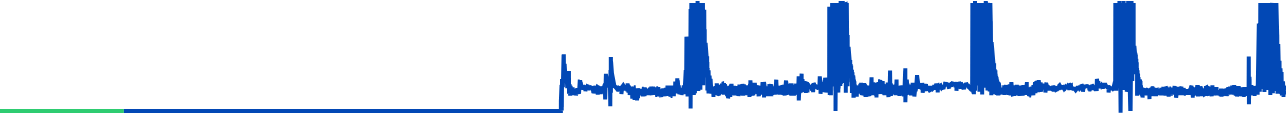}& 
\makecell{Change \\ Point \\ \textbf{(CP)}} & \makecell{\textbf{Exathlon} \\ \texttt{\textbf{id}: 15} \\ \texttt{\textbf{dim}: 6}}  & \makecell{Dynamic\\Allocation}\\
\hline
\centering\includegraphics[width=\linewidth]{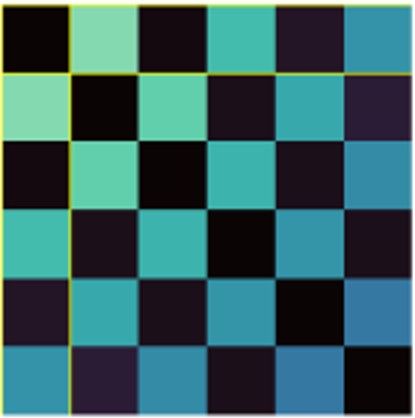} &  
\centering\includegraphics[width=\linewidth]{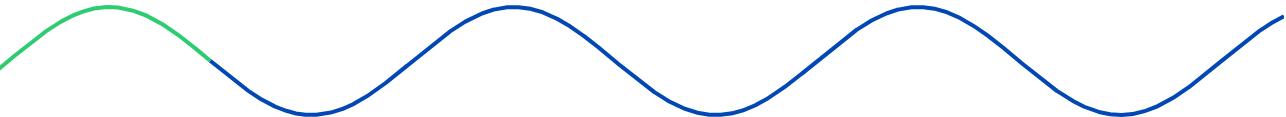}& 
\makecell{Periodic \\ drift \\ \textbf{(P)}} & \makecell{\textbf{CATSv2} \\\texttt{\textbf{id}: 1} \\ \texttt{\textbf{dim}: asin2}} & \makecell{Seasonality} \\
\hline
 \centering\includegraphics[width=\linewidth]{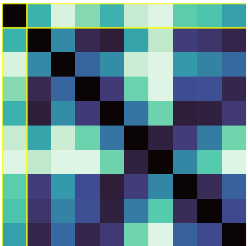}& 
\centering\includegraphics[width=\linewidth]{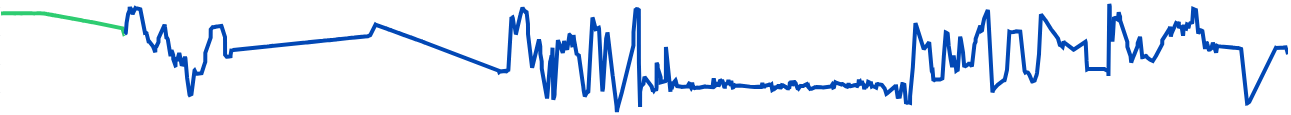} 
& \makecell{Random \\ Walk \\ \textbf{(RW)}} & \makecell{\textbf{OPP.} \\ \texttt{\textbf{id}: 7} \\ \texttt{\textbf{dim}: TAG2X}} & \makecell{Physiological\\Variability}\\
\hline
\centering\includegraphics[width=\linewidth]{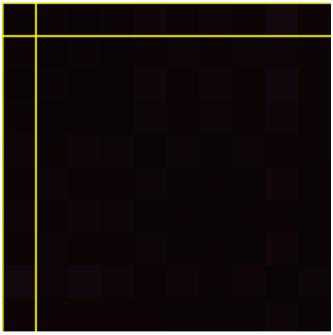} &
\centering\includegraphics[width=\linewidth]{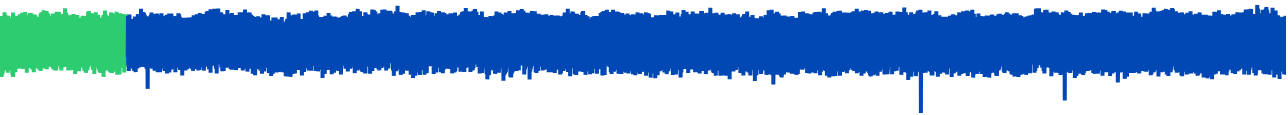}& 
\textcolor{gray}{\textit{No drift}} & \makecell{\textbf{MITDB} \\ \texttt{\textbf{id}: 3} \\ \texttt{\textbf{dim}: V5}} & \makecell{Medical\\(ECG)} \\
\hline
\end{tabular}
}
\vspace{-0.3cm}
\end{table}

\subsection{TSAD methods in \texttt{StrAD}}

\texttt{StrAD} incorporates a diverse set of time-series anomaly detection (TSAD) methods (\emph{Static} and \emph{Online}) and \emph{Streaming} algorithms. This selection enables a systematic comparison across different modeling paradigms, computational profiles and adaptation strategies.

\begin{table}[tb]
\centering
\caption{Static/Online TSAD Methods in \textbf{\texttt{StrAD}} ($^*$: For Online settings, $|T|$ is the Size of the Initial Batch $\mathcal{B}_0$)}
\vspace{-0.3cm}
\label{tab:TSADmethods}
\resizebox{\columnwidth}{!}{
\begin{tabular}{l|ccc}
\hline
\textbf{Acronym} & \textbf{Method} & \textbf{Type} & \textbf{Complexity} \\
\hline
\rowcolor{gray!20}
\multicolumn{4}{c}{\textbf{Distance-based}} \\
\textbf{LOF} & LOF~\cite{LOF}  & Proximity & $\mathcal{O}(|T|\times D)^*$ \\
\textbf{KNN}& $k$-NN~\cite{Hawkins80} &  Proximity & $\mathcal{O}(|T|\times D)^*$ \\
\textbf{KMAD}& $k$-Means~\cite{Hawkins80} & Clustering & $\mathcal{O}(D)$ \\
\textbf{CBLOF} & CBLOF~\cite{CBLOF}  & Clustering & $\mathcal{O}(D)$ \\
\hline
\rowcolor{gray!20}
\multicolumn{4}{c}{\textbf{Density-based}} \\
\textbf{IF} & Isolation Forest~\cite{IForest}  & Tree & $\mathcal{O}(\log(|T|))^*$ \\
\textbf{MCD}& MCD~\cite{MCD}  & Distribution & $\mathcal{O}(D^2)$\\
\textbf{HBOS}& HBOS~\cite{HBOS}  & Distribution & $\mathcal{O}(D)$ \\
\textbf{SVM}& OCSVM~\cite{OCSVM}  & Distribution & $\mathcal{O}(D)$ \\
\textbf{PCA}& PCA~\cite{PCA}  & Encoding & $\mathcal{O}(D)$\\
\textbf{RPCA}& RobustPCA~\cite{RobustPCA}  & Encoding & $\mathcal{O}(D)$\\
\hline
\rowcolor{gray!20}
\multicolumn{4}{c}{\textbf{Prediction-based}} \\
\textbf{CNN}& CNN~\cite{CNN}  & Forecasting & $\mathcal{O}(\ell \times D)$  \\
\textbf{LSTM}& LSTMAD~\cite{LSTMAD}  & Forecasting & $\mathcal{O}(\ell \times D)$ \\
\textbf{AT}& AnomalyTransformer~\cite{AnomalyTransformer}  & Reconstruction & $\mathcal{O}(\ell^2  \times D + \ell  \times D^2)$ \\
\textbf{AE}& AutoEncoder~\cite{AutoEncoder}  & Reconstruction & $\mathcal{O}(\ell \times D)$ \\
\textbf{TrAD} & TranAD~\cite{TranAD}  & Reconstruction & $\mathcal{O}(\ell^2 \times D + \ell  \times D^2)$\\
\textbf{TN} & TimesNet~\cite{TimesNet}  & Reconstruction & $\mathcal{O}(\ell \times \log(\ell)  \times D)$ \\
\textbf{USAD} & USAD~\cite{USAD}  & Reconstruction & $\mathcal{O}(\ell \times D)$\\
\textbf{OA} & OmniAnomaly~\cite{OmniAnomaly}  & Reconstruction & $\mathcal{O}(\ell \times D)$ \\
\textbf{FITS} & FITS~\cite{FITS}  & Reconstruction & $\mathcal{O}(\ell \times \log(\ell) \times D)$ \\
\hline
\end{tabular}
}
\vspace{-0.3cm}
\end{table}

\subsubsection{\textbf{Static/Online TSAD}}

Table~\ref{tab:TSADmethods} shows our representative set of TSAD approaches. The selection balances well-established classical techniques with state-of-the-art deep learning models
to compare diverse modeling paradigms and computational profiles.

More specifically, we include \textit{Proximity-based}   (LOF~\cite{LOF}, KNN\\~\cite{Hawkins80}), \textit{Clustering-based} (KMeansAD~\cite{Hawkins80}, CBLOF~\cite{CBLOF}), \textit{Distribution-based} (MCD~\cite{MCD}, HBOS~\cite{HBOS}, OCSVM~\cite{OCSVM}), \textit{Encoding-based} (PCA\\~\cite{Aggarwal15}, RobustPCA~\cite{RobustPCA}) and \textit{Tree-based} methods (IForest~\cite{IForest}). These approaches remain highly relevant in modern benchmarks~\cite{Liu25, Audibert22, Wagner23}.
Moreover, their relatively low computational overhead make them particularly pertinent in streaming settings. As such, they provide essential reference points for assessing whether more elaborate models yield significant performance improvements.

The \textit{Prediction}-based category 
covers a wide range of methods, 
including Recurrent and Convolutional methods (LSTMAD~\cite{LSTMAD}, CNN~\cite{CNN}), AutoEncoders (AE~\cite{AutoEncoder}, USAD~\cite{USAD}, OmniAnomaly~\cite{OmniAnomaly}), Transformers (Anomaly Transformer~\cite{AnomalyTransformer}, TranAD~\cite{TranAD}) and two Frequency-based encoding (TimesNet~\cite{TimesNet}, FITS~\cite{FITS}).

The complexity described in Table \ref{tab:TSADmethods} refers to the inference complexity (i.e., score computation), where $D$ is the time series dimensionality, $|T|$ the time series length, and $\ell$ the window-size.
Training cost is ignored since all models are trained offline. In streaming settings, inference latency is
the most relevant metric for deployment.

\begin{table}[tb]
\centering
\caption{Streaming TSAD in \textbf{\texttt{StrAD}} ($^*$: Independent to $D$ but Dependent to model hyper-parameters)}
\vspace{-0.3cm}
\label{tab:StreamingMethods}
\resizebox{\columnwidth}{!}{
\begin{tabular}{l|cccc}
\hline
\textbf{Acronym} & \textbf{Method} & \textbf{UM} (Sec~\ref{sec:UM}) & \textbf{MM} (Sec~\ref{sec:MM})& \textbf{Complexity}  \\
\hline
\rowcolor{gray!20}
\multicolumn{5}{c}{\textbf{Numerical}} \\
\textbf{LODA} & LODA~\cite{LODA} & Projections & Tumbling Window & $\mathcal{O}(D)$ \\
\textbf{xS} & xStream~\cite{xStream} & Projections & Tumbling Window & $\mathcal{O}(D)$\\
\textbf{RSH} & RSHash~\cite{RSHash} & Partitioning & Sliding Window (Point) &  $\mathcal{O}(1)^*$\\
\textbf{HST} & HSTree~\cite{HSTree} & Partitioning & Tumbling Window & $\mathcal{O}(1)^*$\\
\textbf{SDOs} & SDOstream~\cite{SDOstream} & Partitioning & Soft Forgetting (Aging) & $\mathcal{O}(D)$\\
\hline
\rowcolor{gray!20}
\multicolumn{5}{c}{\textbf{Structural}} \\
\textbf{RRCF} & RRCF~\cite{RRCF} & Tree & Sliding Window (Point) & $\mathcal{O}(\log (\ell))$\\
\textbf{MCOD} & MCOD~\cite{MCOD} & Clustering & Sliding Window (Point) & $\mathcal{O}(D)$\\
\textbf{LEAP} & LEAP~\cite{LEAP} & Proximity & Sliding Window (Batch) & $\mathcal{O}(D)$\\
\textbf{SKNN} & SWKNN~\cite{SWKNN} & Proximity & Sliding Window (Point) & $\mathcal{O}(D)$\\
\textbf{MemS} & MemStream~\cite{MemStream} & Encoding & Soft Forgetting (Selective) & $\mathcal{O}(D)$\\
\hline
\end{tabular}
}
\vspace{-0.3cm}
\end{table}

\subsubsection{\textbf{Streaming TSAD}}

The streaming methods selected in Table \ref{tab:StreamingMethods} represent the state-of-the-art in streaming TSAD, covering a wide range of adaptation and memory strategies.

The \textit{Numerical} category comprises  
\textit{Projection}-based (LODA~\cite{LODA}, xStr\-eam~\cite{xStream}) and \textit{Partitioning}-based (RSHash~\cite{RSHash}, HSTree~\cite{HSTree}, SDOstream~\cite{SDOstream}). These methods achieve high efficiency by maintaining simple statistics within predefined regions of the data space, making them particularly suitable for streaming scenarios.

The \textit{structural} category features methods with 
dynamically evolving internal representations. 
This group includes \textit{Tree}-based (RRCF\\~\cite{RRCF}), \textit{Clustering}-based (MCOD~\cite{MCOD}), \textit{Proximity}-based (LEAP~\cite{LEAP}, SWKNN~\cite{SWKNN}) and \textit{Encoding}-based mechanisms (MemStream~\cite{MemStream}). 

Finally, our selection also balances different memory management philosophies, ranging from rigid \textit{Tumbling} and \textit{Sliding} windows, to more nuanced forgetting mechanisms.

\subsection{From TSAD to Streaming Settings}

We aim to compare both \textit{Online} and \textit{Streaming} methods in the same benchmark. Thus, we define in this section a common training and normalization procedure.

\subsubsection{\textbf{Initial Batch}}
If needed, models in Table~\ref{tab:TSADmethods} and~\ref{tab:StreamingMethods} are trained on an initial subset of the time series, considered as the \emph{historical} data in a streaming context. The inference of the anomaly score will then be based on the incoming data, which, depending on the nature of the model, can be a point or a sliding window. There is no retraining for \textit{Online} models, as opposed to \textit{Streaming} models, which are based on intrinsic update and/or forgetting mechanism applicable on any incoming data.

\subsubsection{\textbf{Input Normalization}}
 
In this paper, we apply z-score standardization as a preprocessing if needed. The mean and standard deviation are computed \textit{exclusively} from the initial batch and then applied to the incoming stream. The latter is chosen for two reasons: (i) It ensures model agnosticism: local normalization on a per-window basis lack the temporal context to calculate meaningful statistics.  
(ii) It prevents anomalies from skewing the normalization parameters, a common risk in window-based scaling.

 \begin{figure*}[!t]
     \centering
     \includegraphics[width=0.9\textwidth]{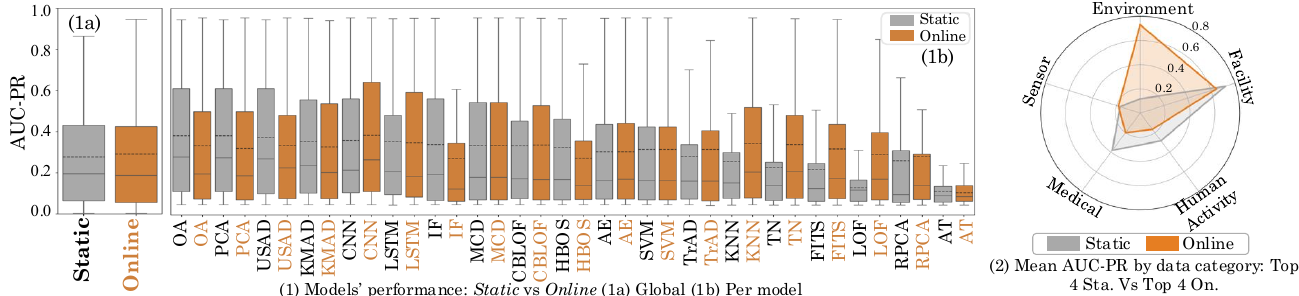}
     \vspace{-0.3cm}
      \caption{\textit{Static} vs. \textit{Online} Accuracy Evaluation on TSB-AD-M: (1) Overall, (2) by Data Category.
     }
     \vspace{-0.3cm}
     \label{fig:static_vs_online}
 \end{figure*}

 \begin{figure}[tb] 
    \centering
    \includegraphics[width=\columnwidth]{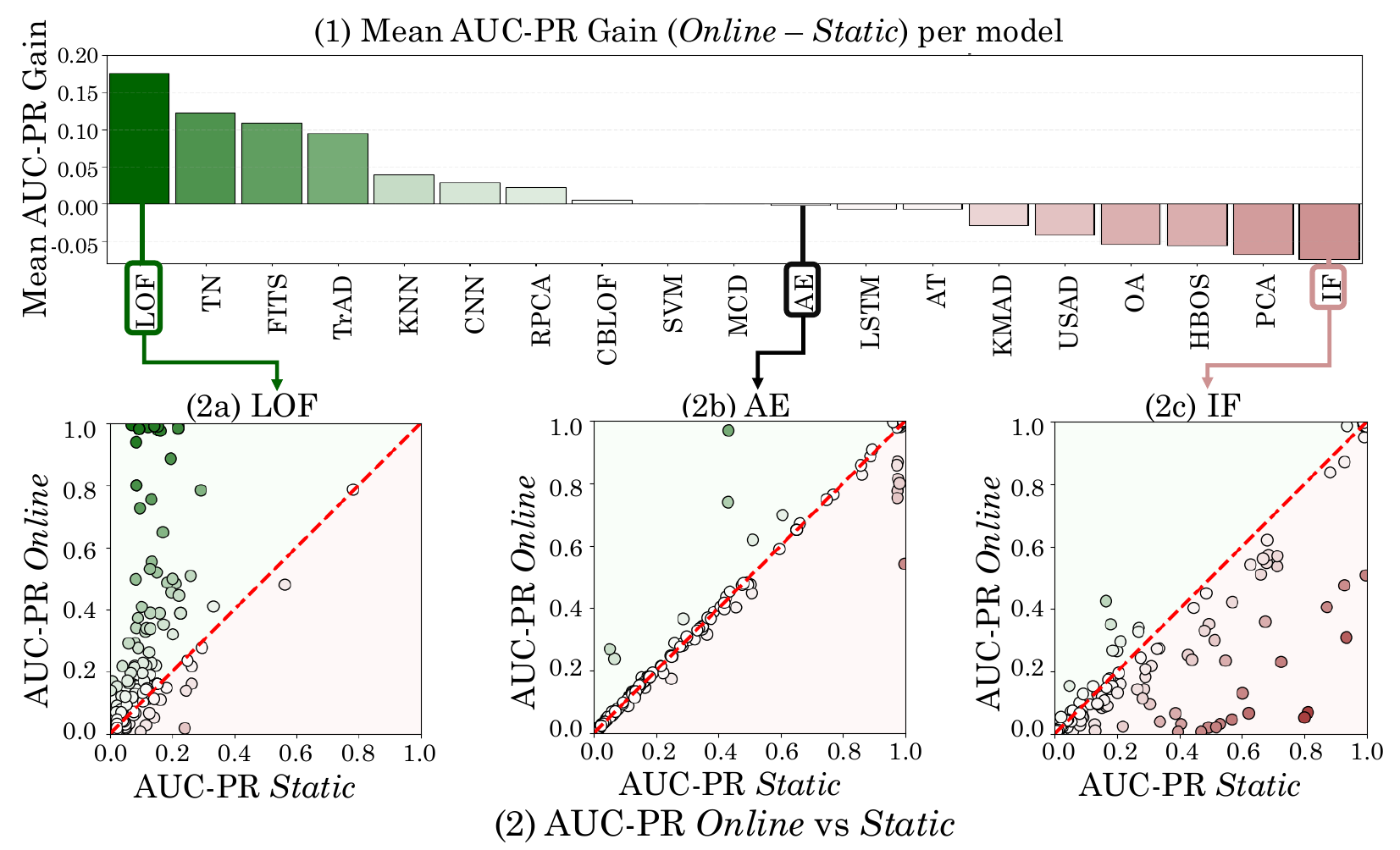}
    \vspace{-0.5cm}
    \caption{(1) AUC-PR gain from Static to Online Settings; (2) Online vs Static on TSB-AD-M for (2a) LOF, (2b) AE, (2c) IF.}
    \label{fig:static_vs_online_detail}
    \vspace{-0.5cm}
\end{figure}

\section{Empirical Evaluation}
\label{experiments}

We now discuss the experimental results addressing the research questions introduced in Section \ref{RQ}.
We provide in Appendix~\ref{Hyperparam} the hyper-parameter calibration. 
All implementations are available in our repository \href{https://github.com/magaliparrino/StrAD}{StrAD}.

\noindent\textit{\textbf{Implementation details.}}
We use the implementation of the \textcolor{blue}{\href{https://thedatumorg.github.io/TSB-AD/}{TSB-AD benchmark} }~\cite{Liu25} for the \textit{Static/Online} methods, in Python. As for the \textit{Streaming} methods, we use MCOD and LEAP's original \textcolor{blue}{\href{https://infolab.usc.edu/Luan/Outlier/CountBasedWindow/DODDS/src/outlierdetection/}{Java implementations}}, and the \textcolor{blue}{\href{https://github.com/CN-TU/dSalmon/tree/master?tab=readme-ov-file}{C++ implementations}} of xStream, SWKNN and SDOstream using the dSalmon~\cite{dSalmon} library. RRCF, HSTree and RSHash are added using the Python library \textcolor{blue}{\href{https://pysad.readthedocs.io/en/latest/}{Pysad} ~\cite{pysad}}. Finally, we adapt MemStream's~\cite{MemStream} 
code to our pipeline.

\noindent\textit{\textbf{Evaluation Measure.}}
The choice of evaluation metric is critical, as many metrics have been shown to exhibit biases~\cite{Liu25, Sorbo24, Wagner23}. This study focuses on parameter-free evaluation measures, such as the Area Under the Curve (AUC) and the Volume Under the Surface (VUS)~\cite{Boniol2025}.
Moreover, as we are not interested in detecting precursors but the anomalies themsleves, and since allowing delay tolerance after the true event contradicts the streaming purpose, VUS’s temporal tolerance is not the most appropriate for this study.
Additionally, the Area Under a Receiver Operating Characteristic Curve (AUC-ROC) metric tends to produce overly optimistic results on highly imbalanced datasets~\cite{Sorbo24}, which is typically the case in anomaly detection. 
Consequently, the Area Under the Precision-Recall Curve (AUC-PR) is preferred in this study.

\subsection{\textit{Static} vs. \textit{Online} TSAD models (Q1)}

 \begin{figure*}[tb] 
     \centering
     \includegraphics[width=\textwidth]{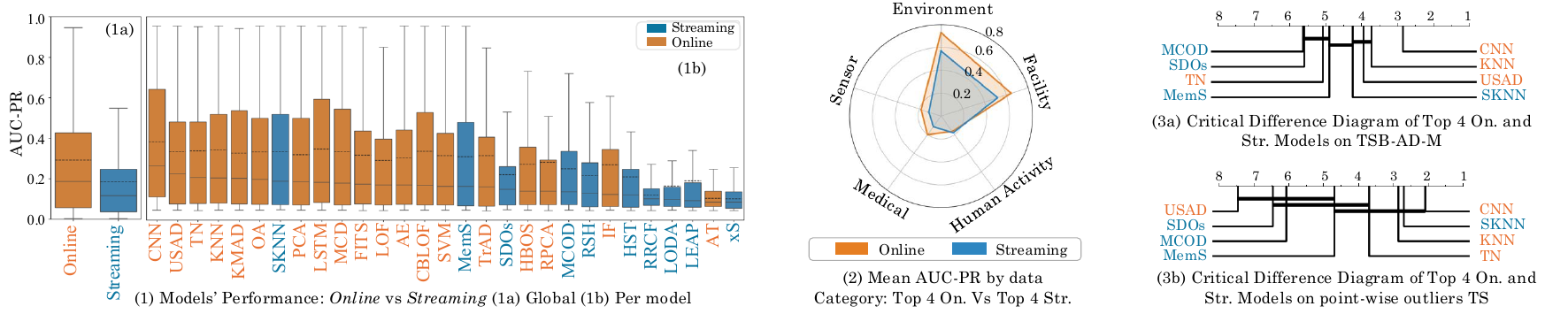}
     \vspace{-0.5cm}
     \caption{\textit{Online} vs. \textit{Streaming} Accuracy on TSB-AD-M: (1) Overall, (2) by Data Category, (3) Critical Difference Diagrams ($\alpha = 0.05$) on (3a) TSB-AD-M and (3b) Point Anomalies Time Series.
     }
     \vspace{-0.3cm}
     \label{fig:online_streaming}
 \end{figure*} 

As shown in Figure~\ref{fig:static_vs_online}, we start by comparing \textit{Static} and \textit{Online} methods.
While the overall performances may appear similar 
in Figure~\ref{fig:static_vs_online} (1a), performances vary per method (Figure~\ref{fig:static_vs_online} (1b)). 
Figure~\ref{fig:static_vs_online} (2) breaks down the mean performance of the four best performing models
in each data category of Table~\ref{tab:datasets}. 
We focus on the top performers to ensure a 'Best-in-Class' comparison, effectively evaluating the theoretical upper bounds of each paradigm. 
We observe strong differences: environment data, which encompasses the TAO dataset, has a large amount of point anomalies throughout the time series. Local anomaly detection, enabled by the online context, better highlights the anomalies, and therefore explains the substantial gap in performance. In contrast, medical data exhibit high regularity and low noise levels. \textit{Static} methods can leverage the entire time series, making them particularly effective for sequence anomalies.

We then measure in Figure \ref{fig:static_vs_online_detail} the performance shifts from \textit{Static} to \textit{Online}. We observe three groups: (i) significant improvement, (ii) stable performance, and (iii) performance drop. 
In the first group, in green, which goes from LOF (Figure \ref{fig:static_vs_online_detail} (2a)) to KNN, we find the \textit{Proximity-based} methods, as well as the two \textit{Prediction-based} methods that use Fourier transform to encode the incoming data. For \textit{Proximity-based} methods, the \textit{Static} environment often suffers from the \emph{multiple similar anomalies} problem, where anomalies mask one another due to their global density. The local context provided by the \textit{Online} environment eliminates this problem, resulting in substantial gains of performance. As for TimesNet and FITS, in \textit{static} environment, the Fourier transform is calculated over the entire time series, meaning the spectral representation can be \emph{polluted} by non-stationarity.
By shifting to an online window, these models better capture the local periodicities, making them more sensitive to temporal deviations that would otherwise be diluted.

Amongst the methods that under-perform in \textit{Online} settings, \textit{Density-based} methods drop the most, such as IForest shown in Figure~\ref{fig:static_vs_online_detail} (2c). In \textit{static} context, these 
models benefit from a comprehensive representation of the data's 
distribution. The \textit{Online} setting \emph{freezes} the distribution of the
training batch as reference. With concept drift, this frozen representation becomes obsolete.

Finally, several models present no significant change. For deep learning-based methods (such as AutoEncoder in Figure~\ref{fig:static_vs_online_detail} (2b)), the transition from \textit{static} to \textit{Online} is almost transparent, as their architecture is already window-based by design.
Similarly, semi-supervised methods, such as OCSVM and MCD, use the training subset to define the boundaries of a \emph{normal} space.  
Thus, \textit{Static} and \textit{Online} settings are essentially the same for these approaches.
\begin{tcolorbox}[colback=gray!5!white,colframe=gray!80!white,boxsep=2pt,top=2pt,bottom=2pt]
\textit{\textbf{Answer to Q1:} Neither static nor online settings are superior; performance depends on the model's underlying logic.}
\end{tcolorbox}

\subsection{\textit{Online} vs. \textit{Streaming} TSAD models (Q2)}
\label{Q2}

We now compare accuracy performances of \textit{Online} methods versus \textit{Streaming} methods in streaming context.
Figure~\ref{fig:online_streaming} (1a) presents counter intuitive results: on average,
\textit{Online} methods significantly outperform \textit{Streaming} methods. 
To ensure this performance gap is not merely due to the initial training batch size, we conduct an ablation study. We evaluate the performance across reduced training sizes of 75\% and 50\% of their original lengths. The results confirm that \textit{Online} dominance is structural rather than data-dependent: \textit{Online} methods maintain a significant lead across all settings, presenting on average a 64\% gain at full training size, 49\% gain at 75\% training size, and 47\% gain even when the training data is cut in half. More details are available in Appendix~\ref{ablation_study}.
This assessment is reinforced when looking at the models' individual performances in Figure~\ref{fig:online_streaming} (1b). Only SWKNN is part of the top half of the benchmark models, and amongst the bottom ten, seven of those methods are \textit{Streaming}. 
This performance disparity remains agnostic of the application domain, as shown in Figure \ref{fig:online_streaming} (2). The top 
performing \textit{Online} methods beat or equal the top 
\textit{Streaming} methods in every domain. 

We propose several hypotheses to explain these findings. Our primary one lies with the architectural target of \textit{Streaming} algorithms. As explained in Section \ref{gaps}, most of them are designed for point outlier detection, meaning they aim to identify isolated instances that fall outside a local density or distance threshold. Therefore, collective anomalies in real-world data are missed. 

\begin{figure}[tb]
    \centering
    \includegraphics[width=\columnwidth]{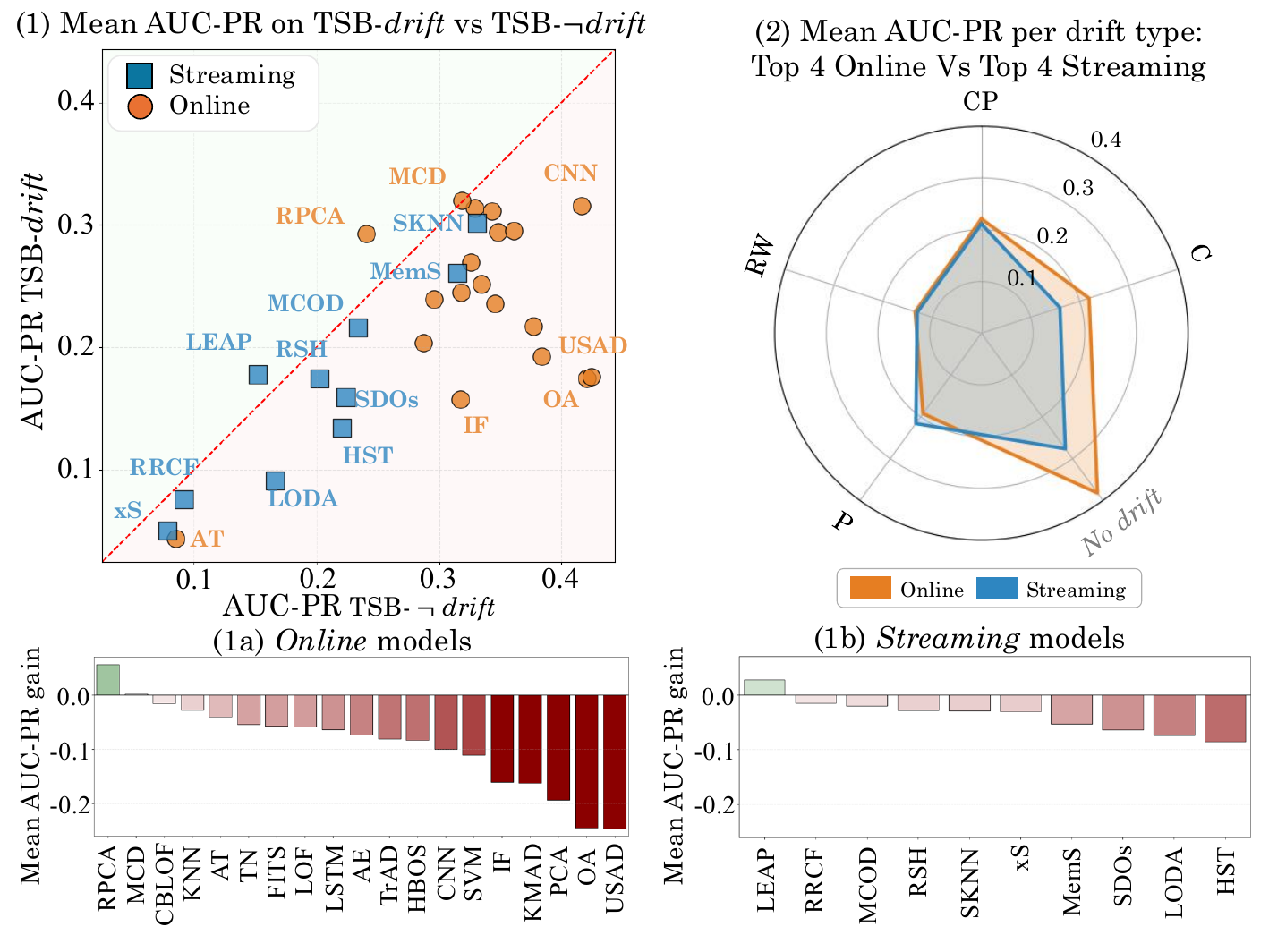}
    \vspace{-0.5cm}
    \caption{(1) TSB-\textit{drift} vs TSB-$\neg$\textit{drift} accuracy; Mean AUC-PR gain of TSB-\textit{drift} on TSB-$\neg$\textit{drift} for (1a) \textit{Online} models and (1b) \textit{Streaming} models; (2) Mean performance on drift-types}
    \label{fig:drift_eval}
    \vspace{-0.3cm}
\end{figure}

To assess the plausibility of this hypothesis, we obtain the critical difference diagrams of the four best models of each category on TSB-AD-M (Figure~\ref{fig:online_streaming}(3a)) and on a subset containing only point anomalies 
(Figure \ref{fig:online_streaming}(3b)). Figure~\ref{fig:online_streaming}(3a) highlights a significant difference of performance between the best \textit{Online} model, CNN, and the best \textit{Streaming} model, SWKNN. Moreover, there are three \textit{Online} models before SWKNN. Conversely, when it comes to detecting point outliers, Figure~\ref{fig:online_streaming}(3b) shows no significant difference between the eight methods. \textit{Streaming} models are more competitive regarding point-wise anomalies than sequences anomalies. 
A second explanation for the difference between \textit{Streaming} and \textit{Online} is the model complexity. \textit{Streaming} methods are by essence designed for efficient update, relying on lightweight, incremental algorithms. Conversely, some \textit{Online} methods are based on high capacity architectures 
which can better capture subtle temporal dependencies.

\begin{figure*}[tb]
    \centering
    \hspace*{-.025\textwidth}\includegraphics[width=1.05\textwidth]{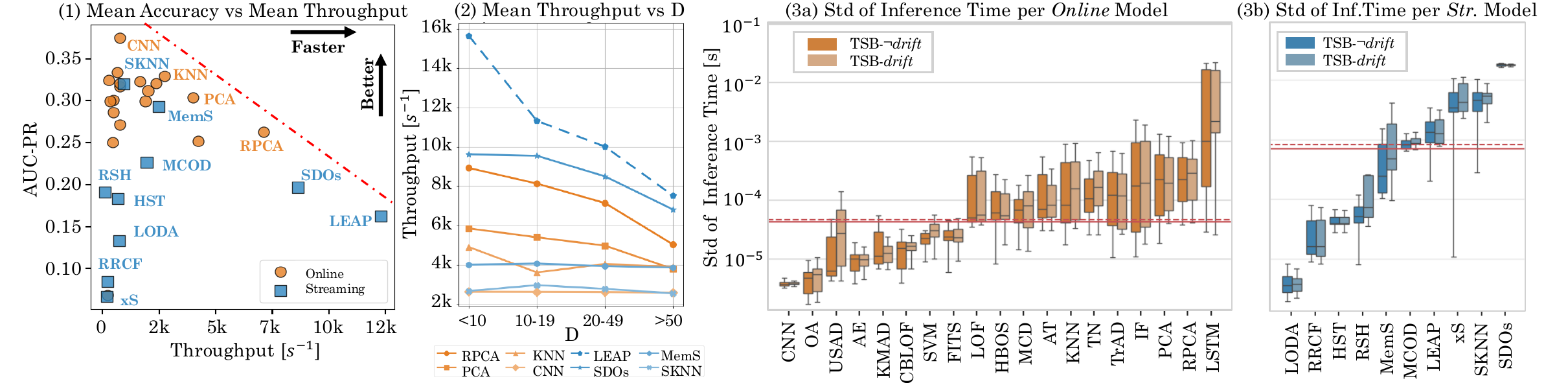}
    \vspace{-0.5cm}
    \caption{ Scalability Evaluation on TSB-AD-M: (1) Accuracy vs Throughput; (2) Throughput vs Number of Dimensions; (3) Inference time standard deviation on TSB-$\neg$\textit{drift} and TSB-\textit{drift}  for (3a) \textit{Online} and (3b) \textit{Streaming} models (Red line highlights median on TSB-$\neg$\textit{drift} and dotted line on TSB-\textit{drift})
    }
    \vspace{-0.3cm}
    \label{fig:time_eval}
\end{figure*}

\begin{tcolorbox}[colback=gray!5!white,colframe=gray!80!white,boxsep=1pt,top=1pt,bottom=1pt]
\textit{\textbf{Answer to Q2:} Our benchmark \texttt{StrAD} shows that streaming approaches are less accurate than online models on real-world data. Streaming methods suffer from a focus on point outliers and struggle to detect temporal nuances.}
\end{tcolorbox}

\subsection{Evaluation on TSB-\textit{drift} (Q3)}

In this section, we evaluate the robustness of \textit{Online} and \textit{Streaming} methods through drifts identified in TSB-\textit{drift}.
Figure~\ref{fig:drift_eval} (1) presents the mean performance of every method on TSB-\textit{drift} compared to TSB-AD-M without TSB-\textit{drift} (written as TSB-$\neg$\textit{drift} in the rest of the paper). Methods below the identity line (highlighted in red) are less performant on TSB-\textit{drift}. 
Nearness to this diagonal can serve as a proxy for drift resilience. In this regard, we observe that a large majority of the \textit{Streaming} methods are amongst the closest to the diagonal. It is further highlighted in Figure \ref{fig:drift_eval} (1a) and (1b), presenting the mean loss when evaluated on TSB-\textit{drift} compared to TSB-$\neg$\textit{drift}. 70\% of the \textit{Streaming} methods (ranging from LEAP to MemStream) show lower performance degradation than 70\% of the \textit{Online} methods (from FITS to USAD). And even the less robust \textit{Streaming} model, HSTree, is less impacted than 7 \textit{Online} models. It is further highlighted in Figure~\ref{fig:drift_eval} (2), showing the mean performance of the top 4 models of each category (CNN, USAD, TimesNet and KNN for \textit{Online} models, and SWKNN, MemStream, SDOstream and MCOD for \textit{Streaming} ones) on the time series presenting each identified type of drift (presented in Section~\ref{Driftcarac}). We can see that in three out of the four types identified (change point, random walk and periodic), there is almost no performance gap. Interestingly, in the case of continuous drift, \textit{Online} methods maintain a lead, though the gap is halved compared to the non-drifted time series. The resilience to continuous drifts for the top four \textit{Online} models compared to their \textit{Streaming} counterparts can be explained by the fact that it is a slow drift. If the models (in particular CNN, USAD and TimesNet) learn a temporal pattern rather than absolute values, they can still recognize the shape despite the drift. 

Nevertheless, this robustness to drift does not translate into absolute superiority. The highest-performing methods remain overwhelmingly \textit{Online}.
The resilience shown by the \textit{Streaming} methods does not bridge the substantial gap highlighted in Section~\ref{Q2}.

\begin{tcolorbox}[colback=gray!5!white,colframe=gray!80!white,boxsep=2pt,top=2pt,bottom=2pt]
\textit{\textbf{Answer to Q3:} While streaming approaches demonstrate superior resilience to concept drifts, they are not necessarily the more appropriate in terms of absolute accuracy.}
\end{tcolorbox}

\subsection{
Efficiency of \textit{Online} vs. \textit{Streaming} (Q4)}

We now evaluate the computational efficiency of \textit{Online} versus \textit{Streaming} methods.
Figure~\ref{fig:time_eval}(1) depicts AUC-PR versus average throughput (average number of inference computations per second) for all time series in TSB-AD. Methods in the top right corner are slower but more accurate, while those in the bottom left corner are the fastest but show the worst accuracy. The top right corner is the ideal sector.
We observe a Pareto frontier, highlighted by the red dotted line, that illustrates a fundamental Performance-Efficiency Trade-off. The \textit{Streaming} models on the Pareto frontier (SDOstream and LEAP) confirm what has been previously underlined: by focusing on high temporal efficiency, those methods have sacrificed TSAD performance on complex data. Conversely, accurate methods are mainly \textit{Online} methods with deep architectures, whose structural complexity limits iteration speed.

Figure~\ref{fig:time_eval}(2) focuses on the Pareto frontier methods, depicting throughput versus the number of dimension.
LEAP is represented by a discontinued line, because it does not return an anomaly score per point, but rather per small batch (of size 32). The bins in Figure~\ref{fig:time_eval}(2) are not strictly equivalent (respectively 58, 45, 54 and 23 time series per bin). As expected, increasing dimensionality negatively impacts the computational efficiency of the methods. Despite this, even the slowest frontier method, CNN, maintains a mean throughput of 770 point scored per second. Thus, this method remains applicable for high-precision monitoring, like electrocardiogram~\cite{MITDB}, or industrial vibration monitoring, that can operate in the 500-800Hz~\cite{vibration_monitoring}. However in high velocity domains, like network intrusion detection~\cite{NID}, high-speed fiber optics links operate at Gbps rates, making such deep models wholly unsuitable.

Finally, we assess the runtime stability in Figure~\ref{fig:time_eval}(3). We show two boxplots per model for the standard deviation of the inference time across TSB-\textit{drift} and TSB-$\neg$\textit{drift}. The red full lines represent the median  
of all models on TSB-$\neg$\textit{drift}, while the red dotted line is on TSB-\textit{drift}. Globally, \textit{Online} methods exhibit higher stability by one order of magnitude. This is expected, as the updating process causes computational instability.
Furthermore, concept drifts slightly exacerbate this instability for \textit{Streaming} methods, as the two medians are almost equal for the \textit{Online} models, but the TSB-\textit{drift} median is distinctly higher for \textit{Streaming} methods in Figure~\ref{fig:time_eval}(3b).

From a practical deployment perspective, these findings outline clear architectural boundaries governed by operational constraints. While streaming methods offer minimal memory footprints and high throughput ideal for resource-constrained contexts or high-velocity data streams, their lightweight nature inherently limits capacity. Conversely, online methods trade higher computational and memory overhead for structural capacity, making them better suited for environments where prediction accuracy on complex, collective anomalies outweighs strict sub-millisecond latency bounds.

\begin{tcolorbox}[colback=gray!5!white,colframe=gray!80!white,boxsep=2pt,top=2pt,bottom=2pt]
\textit{\textbf{Answer to Q4:} While online methods are more accurate, streaming methods dominate the computational high-efficiency end of the spectrum, making them irreplaceable in high velocity domains.}
\end{tcolorbox}

\section{Conclusion}
\label{discussions}

We now conclude on the key findings of
our experimental evaluation on our novel benchmark \texttt{StrAD} and discuss the implications of our work for future research.

\noindent\textbf{Why Do Online Methods Perform Better?}
\textit{Online} methods mitigate anomaly camouflage inherent in \textit{static} settings by limiting temporal context. This helps proximity-based models avoid multiple similar anomalies problem and allows Fourier-based models to focus on dominant frequencies without long-term non-stationarity. Additionally, their higher model capacity enables better temporal dependency modeling than lightweight \textit{Streaming} approaches.

\noindent\textbf{Limitations of Current Streaming TSAD.}
The most significant limitation is the legacy focus on point-wise outliers. Real-world anomalies often consist in collective anomalies, manifesting as pattern changes rather than simple value spikes. Most native \textit{Streaming} algorithms are built on lightweight distance or density metrics designed to identify isolated points and therefore fail to capture such patterns.

\noindent\textbf{Implications for Future Research.}
The observed performance gap indicates that native \textit{Streaming} TSAD still underexplores collective anomalies.
Therefore, \textit{in a streaming world, we should stand still.} But for how long? 
The performance gap identified in our benchmark suggests that native \textit{Streaming} TSAD still has the entire spectrum of collective anomalies left to explore. However, rather than merely simplifying deep models for incremental updates, a more promising frontier may lie
in Streaming Automated Anomaly Detection. While AutoAD~\cite{Bahri22, Liu25autoAD, Sylligardos25} has proven effective in \textit{Static} settings by improving performance and easing model selection for non-experts, its extension to streaming scenarios remains largely unexplored. The latter is a promising research direction for \textit{Streaming} TSAD.

\begin{acks}

\textbf{Funding}
This work was sponsored by the ANRT-CIFRE Grant No. 2025/0400 as part of the PhD thesis of the first author.

\end{acks}


\newpage

\bibliographystyle{ACM-Reference-Format}
\balance
\bibliography{benchmark_online}

\appendix

\section{Hyperparameters Calibration}
\label{Hyperparam}

For reproducibility purpo\-ses, we keep the hyperparameters proposed by the TSB-AD~\cite{Liu25} benchmark for the \textit{static} and \textit{online} configurations. The parameters search ranges for the streaming implementations are determined from previous benchmarks~\cite{Cao24, Vazquez23, Ntroumpogiannis23} and recommendations from the original papers. Grid search is done on the Tuning subset of TSB-AD-M and the tuned hyperparameters are chosen based on the best median performance. 
We present below the search grid in Table~\ref{tab:hyperparam_grid} and tuned hyperparameters obtained for this study in Table~\ref{tab:tuned_hyperparams}.

\begin{table}[h]
\centering
\small
\caption{Hyperparameter Search Grid for the Evaluated Streaming Models}
\label{tab:hyperparam_grid}
\begin{tabular}{lll}
\hline
\textbf{Model} & \textbf{Hyperparameter} & \textbf{Search Range} \\
\hline 
RRCF      & num\_trees & \{2, 4, 8, 16, 20, 25\} \\
          & shingle\_size & \{2, 4, 8\} \\
          & tree\_size & \{256, 512\} \\
\hline
RSHash    & sampling\_points & \{500, 1000\} \\
          & decay & \{0.01, 0.015, 0.02\} \\
          & num\_components & \{50, 100, 200, 300\} \\
          & num\_hash\_fns & \{1, 2, 4\} \\
\hline
HSTree    & window\_size & \{50, 100, 200, 250\} \\
          & num\_trees & \{10, 25, 50, 100\} \\
          & max\_depth & \{10, 15\} \\
\hline
MemStream & memory\_len & \{32, 64, 256, 512, 1024\} \\
          & beta & \{10, 1, 0.1, 0.01, 0.001\} \\
\hline
LEAP      & k & \{5, 10, 20, 40\} \\
          & R & \{0.5, 1.0, 2.0\} \\
          & slidingWindow & \{256, 512\} \\
          & slide & \{32, 64\} \\
\hline
MCOD      & k & \{5, 10, 20, 40\} \\
          & R & \{0.5, 1.0, 2.0\} \\
          & W & \{64, 256, 512, 1024\} \\
\hline
xStream   & window & \{64, 128, 256\} \\
          & n\_estimators & \{50, 100, 200\} \\
          & n\_projections & \{50, 100, 200\} \\
          & depth & \{10, 15, 20\} \\
\hline
SWKNN     & slidingWindow & \{64, 128, 256, 512, 1024\} \\
          & k & \{5, 10, 15, 20, 30, 40\} \\
\hline
SDOstream & k & \{200, 500, 1000, 1500\} \\
          & T & \{128, 256, 512\} \\
          & x & \{3, 6, 10\} \\
\hline
\end{tabular}
\end{table}

\begin{table}[h]
\centering
\small
\caption{Optimal Hyperparameters for the Evaluated Streaming Models}
\label{tab:tuned_hyperparams}
\begin{tabular}{ll}
\hline
\textbf{Model} & \textbf{Tuned hyperparameters} \\
\hline 
RRCF      & \{num\_trees : 16, shingle\_size: 8, tree\_size: 512\} \\
RSHash    & \makecell[l]{\{sampling\_points : 1000, decay: 0.01,\\num\_components: 300, num\_hash\_fns: 2\} } \\
HSTree    & \{window\_size : 200, num\_trees: 25, max\_depth: 15\} \\
MemStream & \{memory\_len : 32, beta: 0.001\} \\
LEAP      & \{k : 40, R: 2.0, slidingWindow: 256, slide: 32\} \\
MCOD      & \{k : 5, R: 1.0, W: 1024\} \\
xStream   & \makecell[l]{\{window : 256, n\_estimators: 200,\\ n\_projections: 50, depth: 20\}} \\
SWKNN     & \{slidingWindow : 64, k: 40\} \\
SDOstream & \{k : 1000, T: 512, x: 10\} \\
\hline
\end{tabular}
\end{table}

\section{More on TSB-\textit{drift}}
\subsection{Synthetic Drift Validation}
\label{synthetic_drift}
We create a synthetic suite, designed to manifest Real Concept Drift (where the relationship $P(Y|X)$ changes) by defining anomalies relative to an evolving local distribution. The base signal is $T_i = \mu_i + \epsilon_i + s_i$ with $\epsilon_i \sim \mathcal{N}(0, 0.1)$ and sparse spikes $s_i$ the labeled anomalies (for $i \in [0, n - 1]$, where $n = |T|$ the length of the time series). We validated our methodology across the following scenarios:

\begin{itemize}
    \item \textbf{Continuous Drift:} Linear mean evolution, $\mu_i = \delta \cdot i, \delta = 3/n$.
    \item \textbf{Change Point:} An abrupt regime shift at midpoint, $\mu_i = 0$ if $i < n/2$ else $2.5$.
    \item \textbf{Periodic Drift:} Toggling distribution parameters every $2 \times t_r$ steps, $\mu_i = 2$ if $\lfloor i / 2t_r \rfloor$ is odd else $0$.
    \item \textbf{Random Walk:} Stochastic non-stationarity, $\mu_i = \sum_{k=1}^{i} \eta_k, \eta_k \sim \mathcal{N}(0, 0.05)$.
    \item \textbf{Virtual Drift (Noise Shift):} To ensure our threshold distinguishes between CD and simple noise increases, we simulated a variance shift: $\sigma = 0.1 \rightarrow 0.4$ at $i = n/2$ ($\mu_i$ constant).
\end{itemize}
 
Plots and codes are available on our repository.

\subsection{Complex Drift Characterisation}
\begin{table}[h]
\centering
\caption{Examples of Complex Patterns Throughout TSB-\textit{drift} (Training Batch $\mathcal{B}_0$ is Highlighted in Green)}
\label{tab:special_drift_matrix_ts}
\resizebox{\columnwidth}{!}{
\begin{tabular}{
|C{2cm}|
C{8.8cm}|
C{2cm}|
C{2cm}|}
\hline
\makecell{\textbf{Matrix} \\ \textbf{Pattern}} & \textbf{Time Series} & \makecell{\textbf{Concept} \\ \textbf{Drift}} & \makecell{\textbf{Dataset} \\ (id TS, d)}\\
\hline
\centering\includegraphics[width=\linewidth]{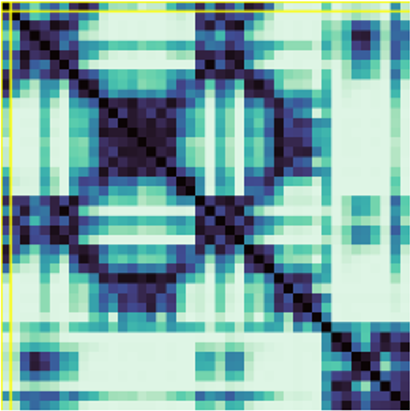} & 
\centering\includegraphics[width=\linewidth]{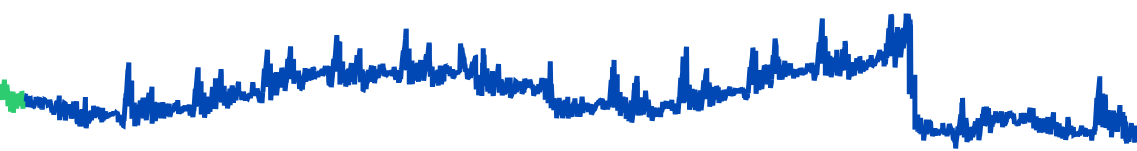}&
\makecell{(P) and \\ (CP)} & \makecell{\textbf{SMD} \\ \texttt{\textbf{id}: 19} \\ \texttt{\textbf{dim}: 5}} \\
\hline
\centering\includegraphics[width=\linewidth]{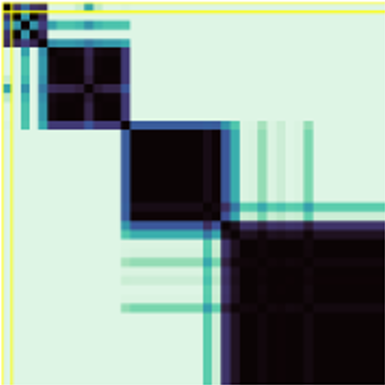}& 
\centering\includegraphics[width=\linewidth]{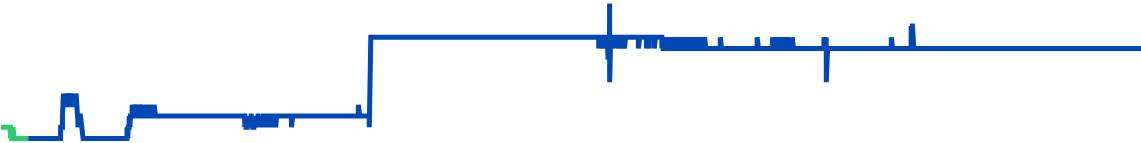}& 
\makecell{Consecutive \\  \textbf{(CP)}} & \makecell{\textbf{SMD} \\ \texttt{\textbf{id}: 19} \\ \texttt{\textbf{dim}: 29}} \\
\hline
\centering\includegraphics[width=\linewidth]{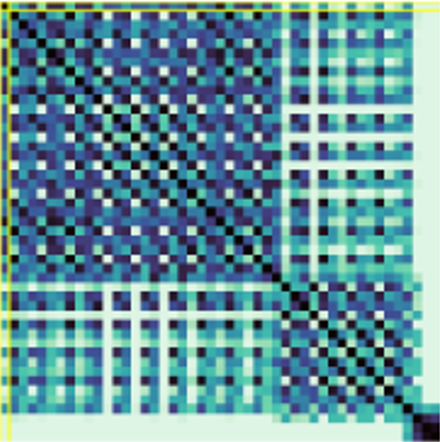} &  
\centering\includegraphics[width=\linewidth]{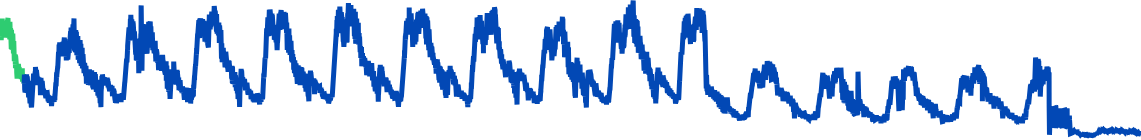}& 
\makecell{Consecutive \\ \textbf{(P)}} & \makecell{\textbf{SMD} \\\texttt{\textbf{id}: 22} \\ \texttt{\textbf{dim}: 18}}  \\

\hline
\end{tabular}
}
\end{table} 

Of the identified patterns illustrated in Table~\ref{tab:drift_matrix_ts}, we can ensue more complex combinations that are still visually identifiable with the matrix patterns such as those identified in Table~\ref{tab:special_drift_matrix_ts}. Among those, we have the re-occurrence of the same pattern, like consecutive Change Points, or a change of periodicity. But we also have the succession of different patterns, such as the sudden interruption of a Periodic pattern.

\section{Ablation Study}
\label{ablation_study}
We conduct this study on time series having at least a 1000 points in the initial training size, as some models cannot have an training batch size lower than 500 points, due to their hyperparameters. The study is therefore conducted on 142 times series, representing 15 out of the 17 datasets (the only time series from CreditCard and the 13 from TAO have an initial training batch size of 500 points and could not be included). The results are summarized in Table~\ref{tab:ablation_training_size};

\begin{table}[h]
\centering
\caption{Ablation Study: Mean AUC-PR Performance Across Different Initial Training Batch Sizes.}
\label{tab:ablation_training_size}
\resizebox{\columnwidth}{!}{%
\begin{tabular}{l@{\qquad}ccc}
\toprule
\makecell{\textbf{Training} \\ \textbf{Size}} & \makecell{\textbf{Mean} \\ \textbf{Online}} & \makecell{\textbf{Mean} \\ \textbf{Streaming}} & \makecell{\textbf{Relative} \\ \textbf{Online Gain (\%)}} \\
\midrule
Full (100\%) & 0.2807 & 0.1713 & +63.91 \\
75\%         & 0.2742 & 0.1847 & +48.51 \\
50\%         & 0.2701 & 0.1836 & +47.10 \\
\bottomrule
\end{tabular}%
}
\end{table}

\end{document}